\documentclass[conference,letterpaper,10pt]{ieeeconf}
\usepackage[left=1.91cm,right=1.91cm,top=1.91cm,bottom=1.91cm]{geometry}

\usepackage{graphicx}
\usepackage{amssymb}
\usepackage{amsmath}
\usepackage{cite}
\usepackage{comment}
\usepackage{color}
\usepackage{url}
\usepackage{alltt}
\usepackage{cleveref}
\usepackage{tikz}
\usepackage{pgfplots}
\usepackage{pgf-umlsd}
\usepackage{cprotect}
\usepackage{multirow}
\usepackage{algorithm}
\usepackage[noend]{algpseudocode}
\usepackage{bigfoot}
\usepackage{paralist}
\usepackage{tabularx}
\usepackage[whole]{bxcjkjatype}
\usepackage{ifthen}
\usepackage{threeparttable}
\usepackage[hang,small,bf]{caption}
\usepackage[subrefformat=parens]{subcaption}
\usepackage[super]{nth}
\usepackage{colortbl}
\usepackage{scalefnt}
\usepackage{threeparttable}

\pgfplotsset{compat=1.14}
\IEEEoverridecommandlockouts
\DeclareMathOperator*{\argmin}{argmin}

\makeatletter
\newcommand{\figcaption}[1]{\def\@captype{figure}\caption{#1}}
\newcommand{\tblcaption}[1]{\def\@captype{table}\caption{#1}}
\makeatother

\title{\LARGE \bf
Target-mass Grasping of Entangled Food\\using Pre-grasping \& Post-grasping
}

\author{
Kuniyuki Takahashi$^{\dagger}$,
Naoki Fukaya$^{\dagger}$,
Avinash Ummadisingu$^{\dagger}$
\thanks{$^{\dagger}$K. Takahashi, N. Fukaya, and A. Ummadisingu are with Preferred Networks, Inc.
        {\tt\footnotesize 
        \{{takahashi, fukaya, ummavi\}@preferred.jp}}}
}

\begin{document}

\maketitle
\thispagestyle{empty}

\begin{abstract}
Food packing industries typically use seasonal ingredients with immense variety that factory workers manually pack. 
For small pieces of food picked by volume or weight that tend to get entangled, stick or clump together, it is difficult to predict how intertwined they are from a visual examination, making it a challenge to grasp the requisite target mass accurately.
Workers rely on a combination of weighing scales and a sequence of complex maneuvers to separate out the food and reach the target mass.
This makes automation of the process a non-trivial affair.
In this study, we propose methods that combines 1) pre-grasping to reduce the degree of the entanglement, 2) post-grasping to adjust the grasped mass using a novel gripper mechanism to carefully discard excess food when the grasped amount is larger than the target mass, and 3) selecting the grasping point to grasp an amount likely to be reasonably higher than target grasping mass with confidence.
We evaluate the methods on a variety of foods that entangle, stick and clump, each of which has a different size, shape, and material properties such as volumetric mass density.
We show significant improvement in grasp accuracy of user-specified target masses using our proposed methods. \footnote{An accompanying video is available at the following link:\\ \url{https://youtu.be/jGYGq5hDybs}}
\end{abstract}

\section{Introduction}
\label{sec:introduction}
In the food packaging industry, the ability to pick up various, highly variable food and place them in their respective positions is essential.
In Japan, ``Bento'' boxes are famous, but there are also boxed lunches in other parts of the world.
Bento combines many types of foods and changes its contents every few weeks, using seasonal ingredients.
Not only do the contents of bento vary greatly, but even multiple pieces of the same kind of food may demonstrate significant differences in shape and size.
Due to the constantly changing menus, automating the process is a major challenge and has only seen minor success with some common foods.
In an era of labor shortages, it has become critical to develop intelligent systems that can learn to adapt to the industry's diversity and scale.

The quantity of food to be placed in a bento box is specified either by the number of pieces for \emph{large pieces of food} (e.g. fried chicken) or by mass for groups of \emph{small pieces of food} (e.g. shredded vegetables).
Prior work in robotics has been successfully demonstrated for the manipulation of \emph{large pieces of food} in the food industry~\cite{matsuoka2020learning, misimi2018robotic, ummadisingu2022luttered}.
However, even though a significant number of \emph{small pieces of food} are used in the industry, their manipulation is relatively unexplored, and their deformable nature and propensity to get entangled, stick and clump makes them difficult to handle.

\begin{figure}[t]
	\centering
	\includegraphics[width=0.70\columnwidth]{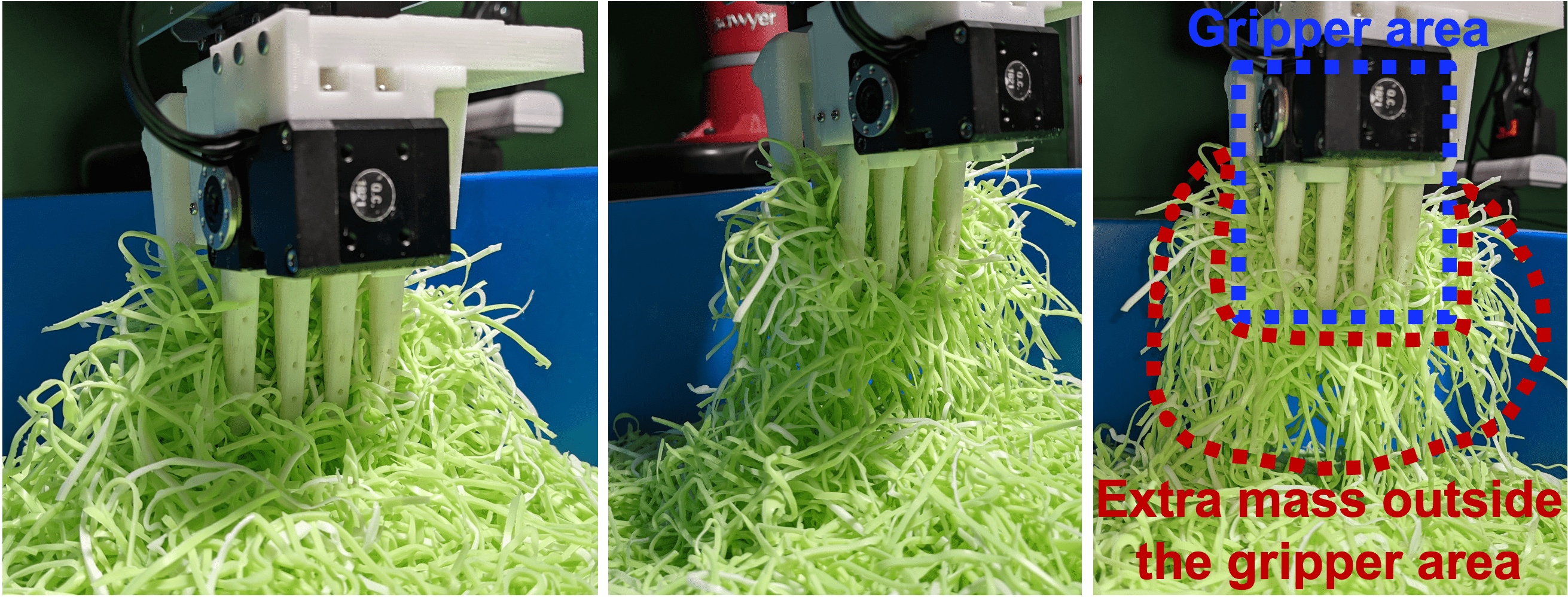}
	\caption{Entangled food grasping: It is difficult to grasp the target mass since the amount of food that is outside of the gripper area when grasped is not constant and unpredictable due to the complex entanglement of pieces each other.}
	\label{fig:difficulty_of_entangle}
\end{figure}

Prior work by us has achieved partial success in the target-mass grasping of granular food that is a part of \emph{small pieces of food} such as peanuts and beans, which is relatively easy to handle by estimating the grasping mass based on RGB and depth information of the food surface~\cite{takahashi2021uncertainty}.
However, foods that may entangled, stick, or clump such as shredded cabbage and bean sprouts, are different from granular foods in that they may intertwine or bunch with each other, and the degree of entanglement is irregular and unpredictable.
This makes it difficult to grasp the desired targeted mass of food since the amount of food that is outside of gripper when grasped is not constant and unpredictable (See Fig.~\ref{fig:difficulty_of_entangle}).
In this paper we refer to these foods as \emph{entangled foods}.

To deal with entangled foods, three things need to be done:
1) reduce the degree of the entanglement of the food, 2) deal with any discrepancies in the estimation of grasping mass caused by entanglement, and
3) the uncertainty of learning from a small dataset and the unpredictable entanglement of foods need to be addressed.
Points 2 and 3 are necessary due to the inherent uncertainty in the grasping process.
In addition to the unpredictability of food entanglement, the partial observability of the food state below the surface, along with uncertainty in sensing and actuation make it a highly stochastic process by nature. 
Additionally, the deformable and entangled nature of the foodstuffs makes simulation unsuitable to generate training data, requiring learning from real-world data.
There is, however, a limit to the amount of data that can be collected due to damage caused by the grasping process, as well as changes caused by temperature, moisture, etc.
This limitation inevitably makes the output values of the trained deep neural networks uncertain.

\begin{figure*}[t]
	\centering
	\includegraphics[width=1.70\columnwidth]{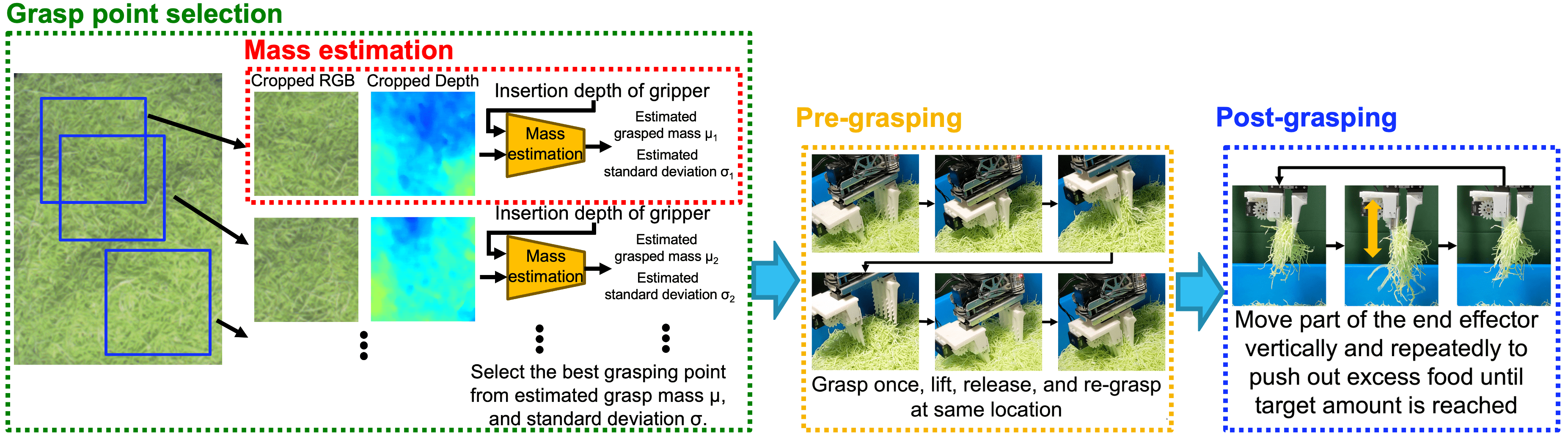}
	\caption{Methods for entangled food grasping: Grasp point selection with mass estimation and predictive  uncertainties to grasp reasonably more than the target mass, pre-grasping to reduce degree of entanglement of foods, and post-grasping to discard the extra grasped mass using a new gripper design and mechanism.}
	\label{fig:method}
	\vspace{-7mm}
\end{figure*}

In this study, we propose the following three methods for grasping the target mass of entangled food: 1) pre-grasping to reduce the degree of entanglement, 2) post-grasping with a new gripper mechanism to adjust the extra mass, and 3) grasp point selection for grasping more than the target mass by estimating the grasping mass from RGB-D images with insertion depth of gripper and considering the uncertainty so it can be adjusted later since accurate mass estimation is difficult due to uncertainty.
\section{Related Work}
\label{sec:related works}
\subsection{Manipulation of Small Pieces of Food}
\label{sec:Manipulation of Small Pieces of Food}
The study of \emph{small pieces of food} has seen limited progress than that of \emph{large pieces of food}, with work focusing primarily on the manipulation of granular foods.
Prior work has looked at the grasping of a fixed target mass of food, which would be difficult to generalize to an arbitrary target mass~\cite{hoerger2019pomdp}.
There has been some research and development on target-mass grasping of granular foods~\cite{clarke2018learning} including our study~\cite{takahashi2021uncertainty}, manipulation for desired shape~\cite{schenck2017learning}, and estimation of food properties~\cite{ matl2020inferring}.
\emph{Entangled foods} behave differently than \emph{granular foods}, in that they are likely to pull along neighbouring pieces when grasped, and predicting the degree of entanglement is difficult.
One study aims to improve the accuracy by reducing entanglement by inserting the gripper into the food and then spreading it out~\cite{ray2020robotic}.
Since it is impossible avoid entanglement entirely, they will inevitably occur and lead to deviations from the target mass.
In this paper, we propose a new method that includes the reduction of entanglement and post-grasping adjustment after an excess amount is grasped.
\subsection{Food Manipulation from the Hardware Perspective}
\label{sec:Food Manipulation from the Hardware Aspect}
Soft-touch grippers are widely used for grasping because they do not damage the food to be grasped and the fingertips are passively adjusted to the food shape, which enables grasping without precise control~\cite{dimeas2015robotic, endo2016development, endo2020robotic, gafer2020quad, dang2021robotic}.
Most of these studies assume that the food to be grasped has a certain size- typically \emph{large pieces of food}, such as bread or strawberries, and are not suitable for grasping \emph{small pieces of food}, such as shredded cabbage or corn.

Attempts have also been made to grasp food by pneumatic adsorption~\cite{sam2010robotic, morales2014robotic, elango2018robotic}.
This method is not suitable for use in food factories where hygiene is essential, because of the accumulation of dirt and foodstuff in the end-effector and piping.
Additionally, in the case of adsorption, if the food to be grasped is small, it may be sucked or blown away, making it challenging to apply.

In the case of such \emph{small pieces of food}, it is mostly desired not only to be able to grasp the food, but also to be able to grasp a certain mass of food in a consistent.
Custom designed grippers, such as scoops of specific sizes, may be used but are highly specific for a food \& target mass and may not be reusable for others.
A gripper capable of grasping a fixed mass of food has been proposed, but it has a problem that the target grasping mass cannot be changed arbitrarily~\cite{wang2021soft}.
Therefore, for practical grasping of \emph{small pieces of food}, it is necessary to be able to grasp a certain mass of food, and have the flexibility to change the target grasping mass according to the situation and requirements.
\section{Method}
\label{sec:method}
Our proposed method to grasp a user--specified target mass of entangled food consists of the following steps for inference (see Fig.~\ref{fig:method}):
A) Mass estimation: Grasp mass predictions along with predictive uncertainties are obtained for RGB-D patches across the food tray with insertion depth of the gripper.
B) Grasp point selection: Select the best grasp point to grasp moderately more than the target grasp mass from estimated grasp mass and uncertainty.
C) Pre-grasping: Reduces the degree of entanglement.
D) Post-grasping: Adjust the grasping amount to the target mass when the grasping amount is more than the target mass.

\subsection{Mass Estimation}
\label{sec:Mass Estimation from RGB-D with insertion depth of gripper}
The mass estimation network predicts the mass of the food to be grasped along with its predictive uncertainties for a cropped patch of the RGB-D image and insertion depth of the gripper.
Uncertainty in this context is the likelihood that the actual grasp mass will be different from the predicted mass for the RGB-D and insertion depth of gripper input.
Predictions with low uncertainty indicate the model's high confidence that a real grasp's outcome will match its prediction.
Ideally, if we identify an insertion depth at a certain patch and order the gripper to grasp it, the gripper should grasp the mass predicted by the network.

The expected grasp mass for a patch $i$ is estimated by the neural network $F(I_{img,i},I_{depth,i}, I_{gripper, i};\xi)=O^{\prime}_{mass,i}$ where the input $I_{img,i}$ and $I_{depth,i}$ are the $i$-th cropped RGB image and depth image, and $I_{gripper,i}$ is insert depth of the gripper, respectively.
$i$ represents the grasp points $(x, y, z)$ in the food tray.
Note that $I_{img}$ and $I_{depth}$ are constant in $z$, and $I_{gripper,i}$ is constant in $x$ and $y$.
$\xi$ is a set of parameters to be optimized.
We process the cropped depth image so that the median of the patch has a value of 0, and the heights of the other points within the patch are relative to it.
The insert depth of the gripper is from the median height of the food surface of the patch.
Note that the opening and closing width can also be given as parameters.
However, as the number of parameters increases, the amount of data required for training increases.
Since it is difficult to collect a large dataset of food as discussed in introduction, we use fixed values.

This task can be cast as a straightforward regression problem where $\xi$ are optimized by minimizing the mean square error (MSE) between the predictions of the network and the training data $MSE(O_{mass,i}, O^{\prime}_{mass,i})$. 
However, since the grasping process is inherently noisy due to numerous factors (including unpredictable entanglement, partial observability, configurations and conditions of the foodstuff, as well as uncertainty in sensing and control), we adopt the Mixture Density Networks (MDN)~\cite{bishop1994mixture}.
Instead of making a point prediction of the expected mass, the MDN is better able to model the underlying process by outputting a class of probability distributions called Mixture of Gaussians or Gaussian Mixture Models.
This allows us to capture uncertainty in predictions with a single forward pass, without the need for multiple models~\cite{lakshminarayanan2017simple}, or the need for Monte Carlo sampling with multiple forward passes~\cite{gal2016dropout} thus making it more suitable for real-time robotics tasks.

The MDN's output is given by 
\begin{equation}
    P(O^{\prime}_{mass,i}) = \sum_{k=1}^{K} \Pi_{k}(I_{i}) \phi\left(O^{\prime}_{mass,i}, \mu_{k}(I_{i}), \sigma_{k}(I_{i})\right)
    \label{eq:fwd_eq}
\end{equation}
where $K$ is a hyperparameter controlling the number of Gaussians, $\Pi_{k}$ is the mixing coefficient for the Gaussian, $\phi$ is the Gaussian function parameterized by the mean $\mu_{k}$ and standard deviation $\sigma_{k}$, both of which are the outputs of the neural network for an input $I_{i}=(I_{img,i},I_{depth,i}, I_{gripper, i};\xi)$

We optimize the parameters of the neural network $\xi$ by minimizing the negative log likelihood of the output distribution given the training data as:
\begin{equation}
    -log\left[\sum_{k=1}^{K} \Pi_{k}(I_{i}) \phi\left(O^{\prime}_{mass,i}, \mu_{k}(I_{i}), \sigma_{k}(I_{i})\right)\right]
    \label{eq:cost_eq}
\end{equation}

\subsection{Grasp Point Selection}
\label{sec:Grasp Point Selection}
In this section, we describe how we select the grasp point from the food tray.
For the mass estimation, we use equation~\eqref{eq:fwd_eq} described in Section~\ref{sec:Mass Estimation from RGB-D with insertion depth of gripper}.
Patches centered around each candidate grasp point are cropped from the RGB-D data and fed into the mass estimation model along with the insertion depth of the gripper. 
However, there is uncertainty in predictions of mass for entangled foods.
Therefore, we attempt to grasp more than the required grasp mass, and then discard the excess through post-grasping movements.
The details of the post-grasping adjustment are described in Section~\ref{sec:Post-grasping}.
If the grasp mass is too much over the target, adjusting the grasp mass takes a long time.
Therefore, it is desired to select the grasp point that does not grasp too much and has low probability of being below the target mass.
To determine how much more to grasp, we use the estimated mean and standard deviation output by the mass estimation model (see equation~\eqref{eq:fwd_eq}).
From the grasp amount $\mu_{i}$ and standard deviation $\sigma_{i}$ estimated from each patch of the food tray, we select the grasping point $i=(x, y, z)$ that satisfies the following:
\begin{equation}
    \argmin_{i} (| M_{tm} - \mu_{i} | + \sigma_{i}), \quad s.t. \quad M_{tm} + \alpha\sigma_{i} < \mu_{i}
    \label{eq:grasp_point}
\end{equation}
where $M_{tm}$ is target mass, and $\alpha$ is a coefficient.
The magnitude of $\alpha$ determines how much more the robot grasps than the target mass, and how likely it is that the robot will be able to acquire more than the target mass.

Therefore, by setting the grasp position to where the estimated grasp mass is greater than or equal to $M_{tm}+\alpha\sigma$, we can express the grasp position that likely achieves a grasp mass greater than the target grasp mass.
However, adjusting the $\alpha$ parameter presents a trade-off, because if $\alpha$ is increased, it is more likely to grasp more than the target grasping mass, but the quantity to be grasped likely increases and it takes additional time to discard the excess.
Since the upper limit of the grasping mass is constrained by the size of the gripper, if $\alpha$ is too large, there will be no grasping position that satisfies the condition.
The standard deviation $\sigma$ serves as a good measure of predictive uncertainty for a given input.
The larger this value estimated by the model, the larger the uncertainty.
Therefore, by selecting a grasp that satisfies the equation~\eqref{eq:grasp_point}, the point with the smallest uncertainty with difference between the target mass and estimated grasping mass is selected.
By incorporating uncertainty into the decision process, we increase the likelihood of grasping an amount greater than the target. Furthermore, by working off the principle that we only need one high confidence grasp candidate, we are able to grasp reliably despite learning from very small datasets as shown in prior work~\cite{takahashi2021uncertainty}.
\begin{figure}[tb]
	\centering
	\includegraphics[width=0.90\columnwidth]{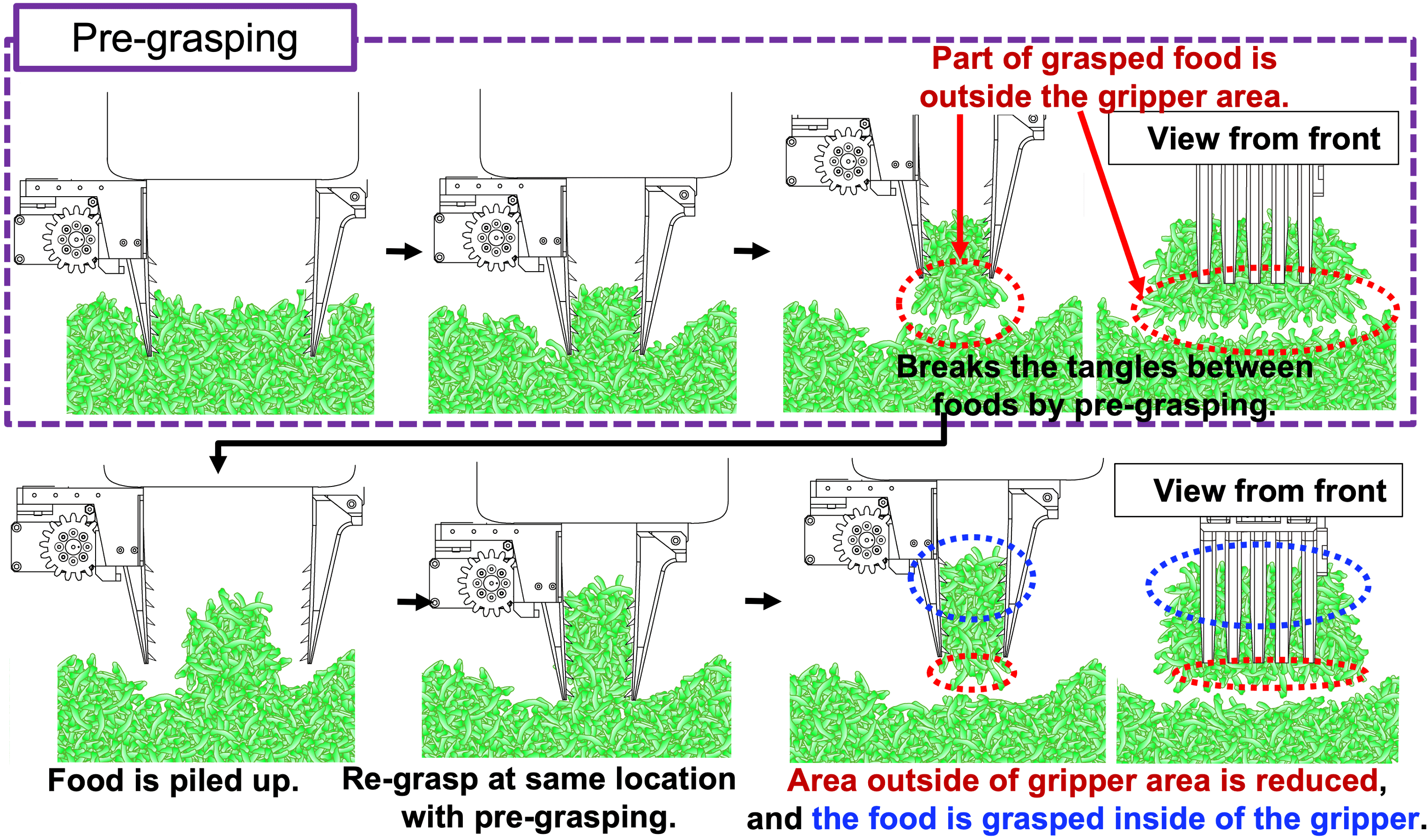}
	\caption{Concept of pre-grasping.}
	\label{fig:pregrasping}
\end{figure}
\subsection{Pre-grasping}
\label{sec:Pre-grasping}
This section describes the purpose of pre-grasping and its methods.
The purpose of pre-grasping is to reduce the degree of the entanglement of foods so that the amount of grasping can be adjusted during post-grasping.
The pre-grasping method is performed when the food at a randomly selected grasping position during data collection or at a grasping position selected by the grasp point selection during inference.
The pre-grasping motion is lifting the food upward, and releasing it.
Then, the food is grasped again at the same grasping position (see Fig.~\ref{fig:pregrasping}).
This sequence of actions aims to achieve the produce the following effects:
1) In grasping and lifting the food, we are able to separate and reduce the degree of the entanglement of the food (See the right-most part of Fig.~\ref{fig:difficulty_of_entangle} and the top most-right part of Fig.~\ref{fig:pregrasping}).
2) Once the food is lifted and released, the overall height of the food becomes larger than before pre-grasping (See the bottom most-left part of Fig.~\ref{fig:pregrasping}).
As a result after pre-grasping, food present outside of gripper area is reduced, and the more food is inside the gripper area (See right-most top and bottom of Fig.~\ref{fig:pregrasping}).
If pre-grasping is not performed, the food is grasped with the food just outside the fingertips of the gripper continuing to be entangled.
Therefore, the extra food portion falls as one lump during post-grasping, and the grasping amount cannot be adjusted finely.
The results of this experiment are shown in Section~\ref{sec:Results of Pre-grasping}.
\begin{figure}[t]
	\centering
	\includegraphics[width=0.80\columnwidth]{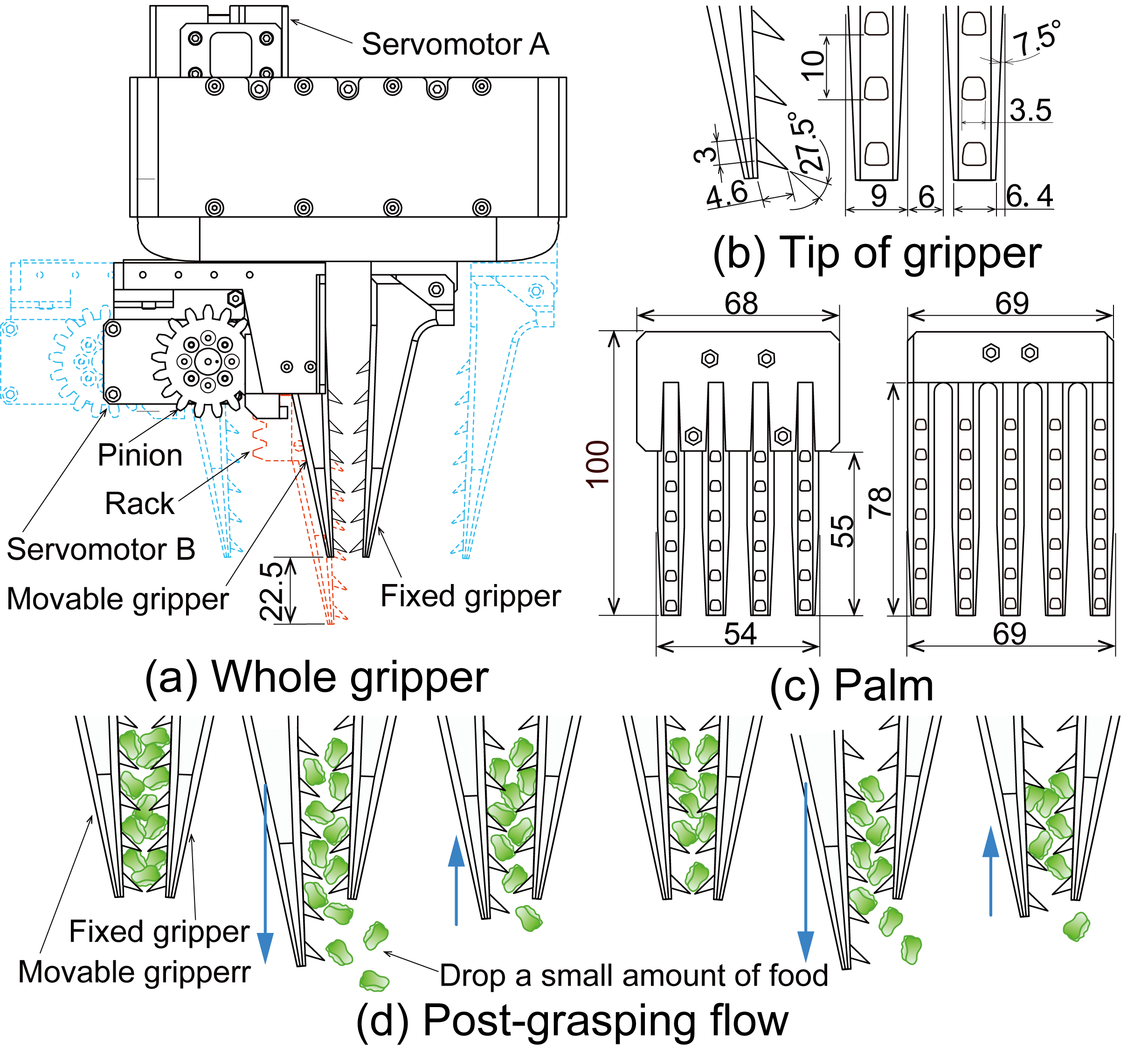}
	\caption{Post-grasping gripper and how it works}
	\label{fig:postgrasp_gripper}
\end{figure}

\subsection{Post-grasping}
\label{sec:Post-grasping}
This section describes the mechanism of the new gripper and its control method to adjust the grasping amount to achieve the target grasping mass when the grasping mass is more than the target mass.
In the post-grasping phase, it is required to achieve an arbitrary target grasping mass with high accuracy.
Since the foods we are interested in are small pieces of food that tend to entangle, stick, or clump, it is difficult to adjust the amount of grasping because the foods lump together and fall out in clumps when discarded making fine-grained control difficult.

It is necessary to drop a small amount of food at a time while simultaneously separating it and overcoming the entanglement.
To achieve this, we developed a new gripper with a novel mechanism that pushes food downward by operating the gripper in a vertical motion (see Fig.~\ref{fig:postgrasp_gripper}~(a)) and a mechanism with spines on the gripper surface (see Fig.~\ref{fig:postgrasp_gripper}~(b) and (c)).
As the gripper surface is covered with spines, the grasped food is gradually pushed downward and falls from the gripper as the movable gripper moves up and down (see Fig.~\ref{fig:postgrasp_gripper}~(d)).

The hardware of the gripper is described below.
The gripper is operated as a parallel gripper by a servomotor A (Dynamixel XM430-W350-R).
The movable gripper, which moves up and down, consists of four forks and is moved up and down by a servomotor B (Dynamixel XM430-W210-R) via rack and pinion with a width of 22.5~mm (see Fig.~\ref{fig:postgrasp_gripper}~(a)).
As shown in Fig.~\ref{fig:postgrasp_gripper}~(c), the movable gripper has four forks and the fixed gripper has five forks.
The forks move in the grooves of the opposing forks.
Therefore, even if the movable gripper moves up and down during the post-grasping motion, spines do not interfere to the opposing forks and can move smoothly.

The faster the speed at which the movable gripper is moved up and down, the more the food that falls from the gripper per unit time, and thus the time required for achieving the target mass is reduced.
However, there is a possibility that the amount of food dropped is too large and falls below the acceptable deviation from the target grasping mass.
For this reason, the motion speed is adjusted according to the target grasping mass and the current grasping amount.
For simplicity, we set the minimum and maximum speed of the gripper and reduce speed linearly while controlling the current speed of the gripper between the the target grasping mass and current grasped mass.
The grasped mass is continuously updated with the help of the scales constantly measuring the discarded mass.
\begin{figure}[tb]
	\centering
	\includegraphics[width=0.75\columnwidth]{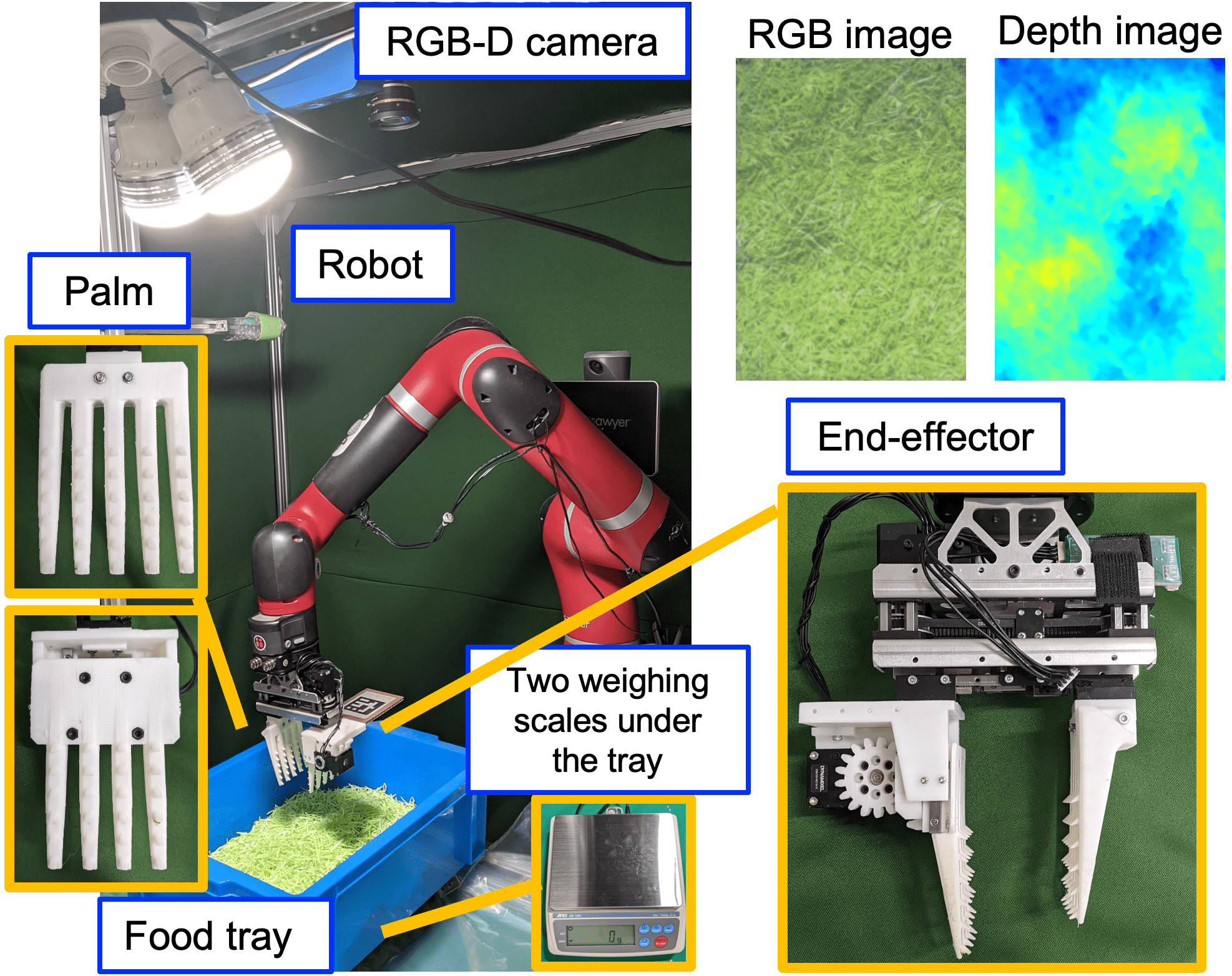}
	\caption{Experiment setup: Robot setup used for the experiment and an example RGB and depth images}
	\label{fig:setup}
\end{figure}

\section{Experimental Setup for Food Picking}
\label{sec:experimental setup}
\subsection{Robot Setup for Data Collection System}
\label{sec:Robot Setup for Data Collections System}
\begin{figure}[tb]
	\centering
	\includegraphics[width=1.00\columnwidth]{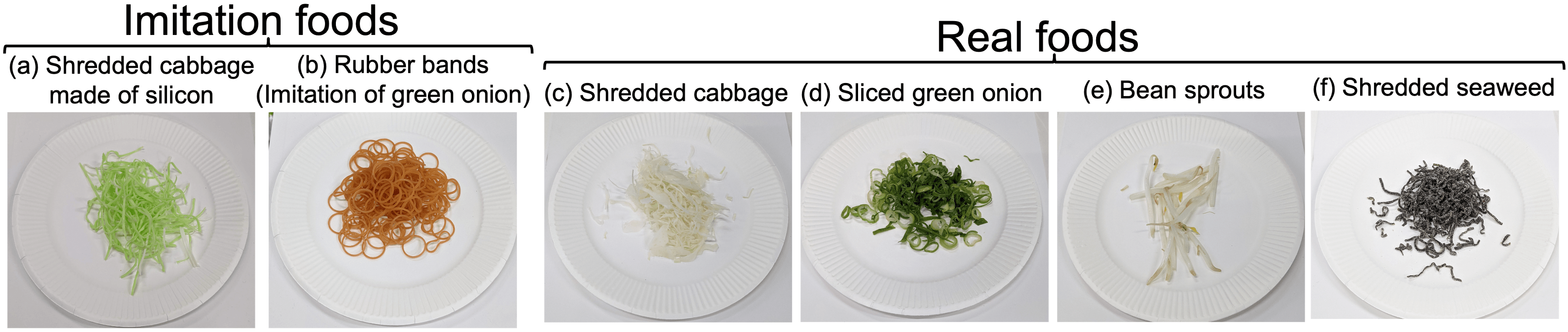}
	\caption{Imitation and real foods used for experiments. All foods in the image are 10g.}
	\label{fig:foods}
\end{figure}

Our robotic system, shown in Fig.~\ref{fig:setup}, consists of a Sawyer 7-DoF robotic arm equipped with proposed post-grasping gripper as an end-effector (see Fig.~\ref{fig:postgrasp_gripper}).
Furthermore, we use an Ensenso N35 stereo camera in combination with an IDS uEye RGB camera to overlook the workspace of the robot arm and use them to retrieve registered point clouds of the scene.
The depth information of each point cloud is acquired at an interval of about 1~mm in the XY axis direction and the accuracy in the depth direction is about 1.5~mm.
We use a food tray and the inner dimensions of the food tray are 424~mm${\times}$308~mm${\times}$160~mm.
Two EK-3000i weighing scales are placed under the food tray, and the amount grasped by the robot is calculated from the mass of the tray before, during, and after the robot picks and post-grasping.
The resolution of the weighing scale is 0.1~g.
The Sawyer, gripper, RGB-D sensor, and weighing scales are connected to a PC running Ubuntu 16.04 with ROS Kinetic.
We target two imitation foods, (a) shredded cabbage made of silicon and (b) rubber bands, and four real foods, (c) shredded cabbage, (d) green onion, (e) bean sprouts, and (f) shredded seaweed, which have different shape, softness, friction and densities (Fig.~\ref{fig:foods}).

\algdef{SE}[DOWHILE]{Do}{doWhile}{\algorithmicdo}[1]{\algorithmicwhile\ #1}%
\begin{algorithm}[t]
    \caption{Data Collection Process}
    \label{alg:datacollection}
    \begingroup
        \scalefont{0.70}
        \begin{algorithmic}[1]
        \For{$\rm{\textsc{iter}}=1$ to $N$}
        	\State $q_{angles} \gets \textsc{home}$ \Comment{Move out of view.}
            \State $RGBD \gets$ CaptureImage\&Depth()
            
            \While{Depth.ContainsNaN()}
                \State RunningAverage() 
            \EndWhile
            \State Process(RGBD)
            \State $X, Y, Z \gets $ GenerateRandomXYCoordinates()
            \State Pre-grasping(X, Y, Z)
            \State Pick(X, Y, Z)
        	\State MeasureScales() \Comment{Measure picked mass}
            \State Place()
        \EndFor
        \end{algorithmic}
    \endgroup
\end{algorithm}
\algdef{SE}[DOWHILE]{Do}{doWhile}{\algorithmicdo}[2]{\algorithmicwhile\ #2}%
\begin{algorithm}[t]
    \caption{Inference Process}
    \label{alg:inference}
    \begingroup
        \scalefont{0.70}
        \begin{algorithmic}[1]
        \For{$\rm{\textsc{iter}}=1$ to $N$}
        	\State $q_{angles} \gets $home \Comment{Move out of view.}
            \State $RGBD \gets $ CaptureImage\&Depth()
            
            \While{Depth.ContainsNaN()}
                \State RunningAverage() 
            \EndWhile
            \State Process(RGBD)
            \State $X, Y, Z \gets $GraspPointSelection()
            \State Pre-grasping(X, Y, Z)
            \State Pick(X, Y, Z)
        	\State MeasureScales() \Comment{Measure picked mass}
        	\If{Grasp mass $\leq$ Target mass-2.0~g}
        	    \State Continue() \Comment{Release and retry from line2}
        	\EndIf
        	\While{Target mass+2.0~g $\leq$ Grasp mass}
                \State Post-grasping()
                \State MeasureScales()
            \EndWhile
            \State Place()
        \EndFor
        \end{algorithmic}
    \endgroup
\end{algorithm}
\subsection{Data Collection Process}
\label{sec:Data Collection Process}
We built a data collection system to collect data without the need for human supervision or labeling effort.
Algorithm (ALGO)~\ref{alg:datacollection} details the data collection process.

We choose the home position such that the robot is out of view of the food tray.
The system captures an RGB-D image and fills in all NaN values, as seen in lines 4-5 of ALGO~\ref{alg:datacollection}.
In line 6 of ALGO~\ref{alg:datacollection}, we process the RGB-D images in the following ways.
From the entire RGB-D image, we obtain cropped patches of $160{\times}160$ pixels with the grasp point as the center of the patch.
1 pixel corresponds to roughly 1~mm of the tray.
The depth information is used to compute the relative heights from the food surface's median height at the cropped patches.
Therefore, the system will work even if the amount of food in the container is gradually reduced by picking from the food tray and placing outside the container.
Note that this is an arbitrary choice to normalize input height.
Then it was converted to RGB by OpenCV's pseudo-coloring API, applyColorMap()~\cite{bradski2008learning}.
Note that in line 7, the pick and place poses are randomly selected within the tray areas, which are located through Aruco marker~\cite{garrido2016generation, romero2018speeded}.
The $x$, $y$, and $z$ coordinates are randomized.
For data collection, the value of $z$ is randomly selected from three values.
Note that the value of $z$ is selected such that end-effector does not collide with the bottom of the food tray.
For imitation cabbage, rubber bands, real cabbage, and green onion, the values of $z$ are selected from [2.0~$\mathrm{cm}$, 3.0~$\mathrm{cm}$, 4.0~$\mathrm{cm}$].
For bean sprouts and seaweed, the values of $z$ are selected from [1.0~$\mathrm{cm}$, 1.5~$\mathrm{cm}$, 2.0~$\mathrm{cm}$] because the robot could not grasp well due to the hardness of the food when the insertion depth of the gripper was deep.
After learning with these three values, the generalization performance allows inference with other values as well.
The details of the inference process are described in Section~\ref{sec:Inference Process}.
The surface of the food in tray continue to evolve significantly and constantly (See Fig.~\ref{fig:setup}).
This allows us to capture diverse surface configurations that may arise in a real factory setting.
In line 8-9 of ALGO~\ref{alg:datacollection}, the robot performs pre-grasping, lifting the food upward, and releasing it.
Then, the food is grasped again at the same grasping position.

The whole pick-and-place process, including data storage, takes about 11.0 to 13.0 sec per cycle (See attached video).
For the training data, we collected 200 data points for each food, respectively.
Out of 200 samples collected, 150 were used for training, and 50 were used to evaluate the network model.
During training of the network(s), we improve generalization by exploiting the gripper symmetry and apply vertical and horizontal flipping of input RGB-D independently with a probability of 0.5.
In addition, we randomly crop patches of $150{\times}150$ pixels from the original $160{\times}160$ to use for training.
\begin{figure}[tb]
	\centering
	\includegraphics[width=0.80\columnwidth]{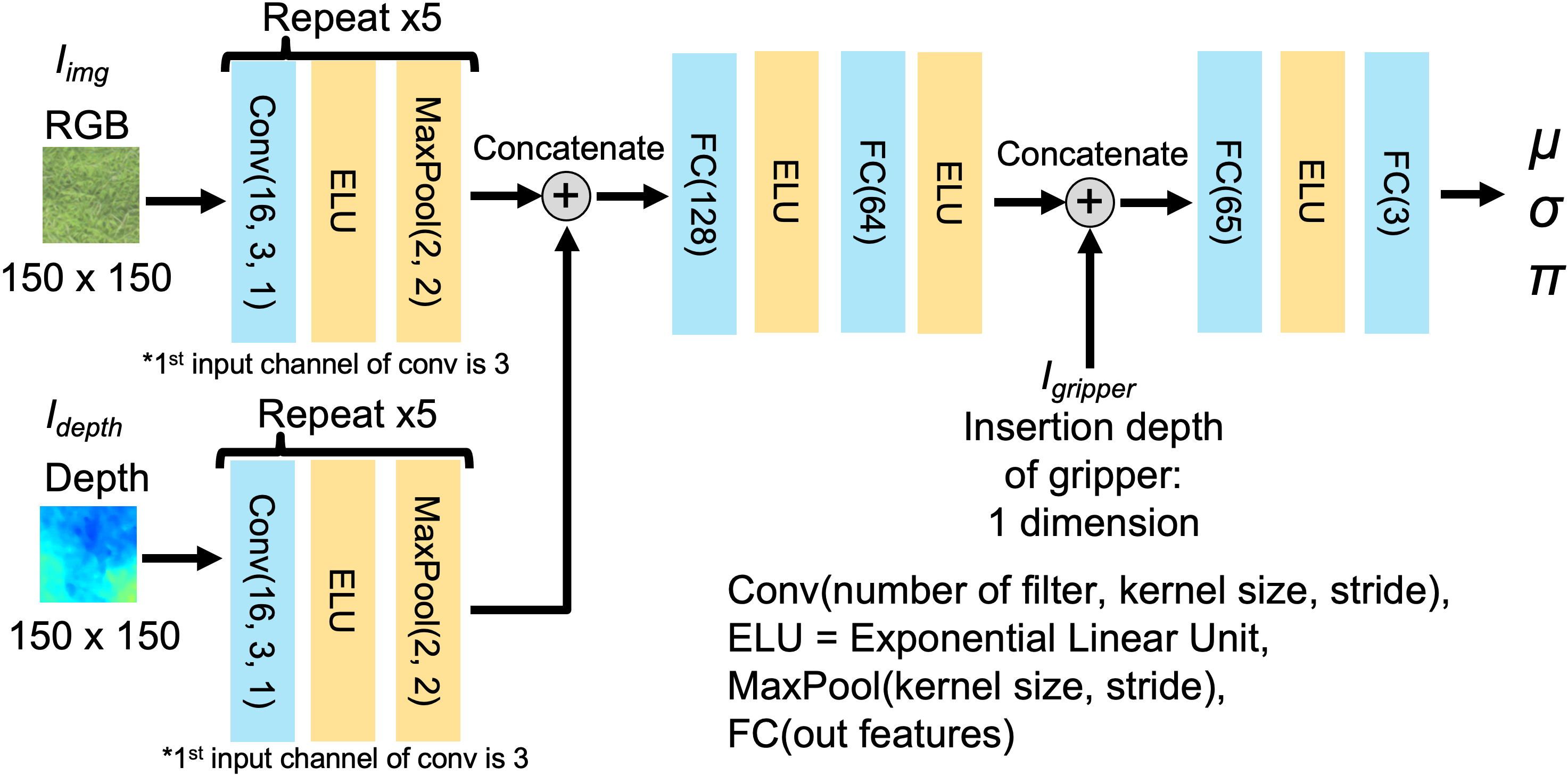}
	\caption{Network design of mass estimation}
	\label{fig:network}
\end{figure}

\subsection{Inference Process}
\label{sec:Inference Process}
In this section, we explain the inference process using the trained model from data collected as described in the previous section.
The process is similar to the data collection process, but differs in two key ways:
1) The grasping position is selected by grasp point selection (line 7 of ALGO~\ref{alg:inference}).
2) The post-grasping is performed until the grasping mass becomes less than (target mass+2.0~g) (line 13-15 of ALGO~\ref{alg:inference}).

In this experiment, the grasp point selection is a grid search of width 15 pixels, which corresponds to about 1.5~$\mathrm{cm}$ in the XY direction in the graspable area of the food tray.
As for the Z-direction, values of [2.0~$\mathrm{cm}$, 2.25~$\mathrm{cm}$, 2.5~$\mathrm{cm}$, 2.75~$\mathrm{cm}$, 3.0~$\mathrm{cm}$, 3.25~$\mathrm{cm}$, 3.5~$\mathrm{cm}$, 3.75~$\mathrm{cm}$, 4.0~$\mathrm{cm}$] are evaluated for imitation cabbage, rubber bands, real cabbage, and green onion.
For bean sprouts and seaweed, values of [1.0~$\mathrm{cm}$, 1.25~$\mathrm{cm}$, 1.5~$\mathrm{cm}$, 1.75~$\mathrm{cm}$, 2.0~$\mathrm{cm}$] are evaluated because the robot could not grasp well due to the hardness of the food preventing insertion of the gripper to deeper levels.
It is possible that a better grasping point could be found if the grid width were made finer but it comes at additional computational overhead.
Inference takes an additional 1.5 sec over the cycle time reported in Section~\ref{sec:Data Collection Process}.
From these evaluated candidates, the best grasping position is selected according to equation~\eqref{eq:grasp_point}.
Note that the grasping position is selected such that end-effector does not collide with the bottom of the food tray by setting predictions at those regions to a grasp amount of 0 and standard deviation of inf.
If the grasped mass at the selected grasp position is less than (target grasp amount-2.0~g), the grasped food is released and the process starts over from line 2 of ALGO~\ref{alg:inference}.

If the grasped mass is greater than (target mass+2.0~g), post-grasping is performed until the mass is less than (target grasping mass+2.0~g).
The mass of food grasped by the robot is measured by weighing scales under the food tray.
The update rate of the weighing scale is 10~$Hz$, which is slower than the control cycle of the post-grasping gripper, about 30~$Hz$.
Furthermore, there is a lag time between when the food leaves the gripper by post-grasping and the time it falls onto the food tray.
In addition, since the falling food exerts a impulse force on the weighing scales, the value of the weighing scale instantaneously measures a value greater than that of the actual falling food.
For these reasons, it is difficult to accurately and quickly measure the mass of grasping during post-grasping, and in this experiment, the accuracy was limited to within (target mass$\pm$2~g).
The introduction of a force sensor on the robot's wrist instead of weighing scales does not improve accuracy.
This is because the gripper is always in motion due to post-grasping, and the value of the force sensor is more unstable than the weighing scales.
\subsection{Deep Learning \& Training \& Inference}
\label{sec:Deep Learning}
We trained the networks on a machine equipped with 256\,GB RAM, an Intel Xeon E5-2667v4 CPU, and Tesla P100-PCIE with 12GB.
Training for the model (initialized randomly for each food) varied from 3 to 15 minutes for each food.
Details of the network architectures is shown in Fig.~\ref{fig:network}.
Our experiments for inference with trained model were performed on a machine equipped with 31.3\,GB RAM, an Intel Core i7-7700K CPU, and GeForce GTX 1080 Ti.

\begin{figure}[tb]
	\centering
	\includegraphics[width=0.90\columnwidth]{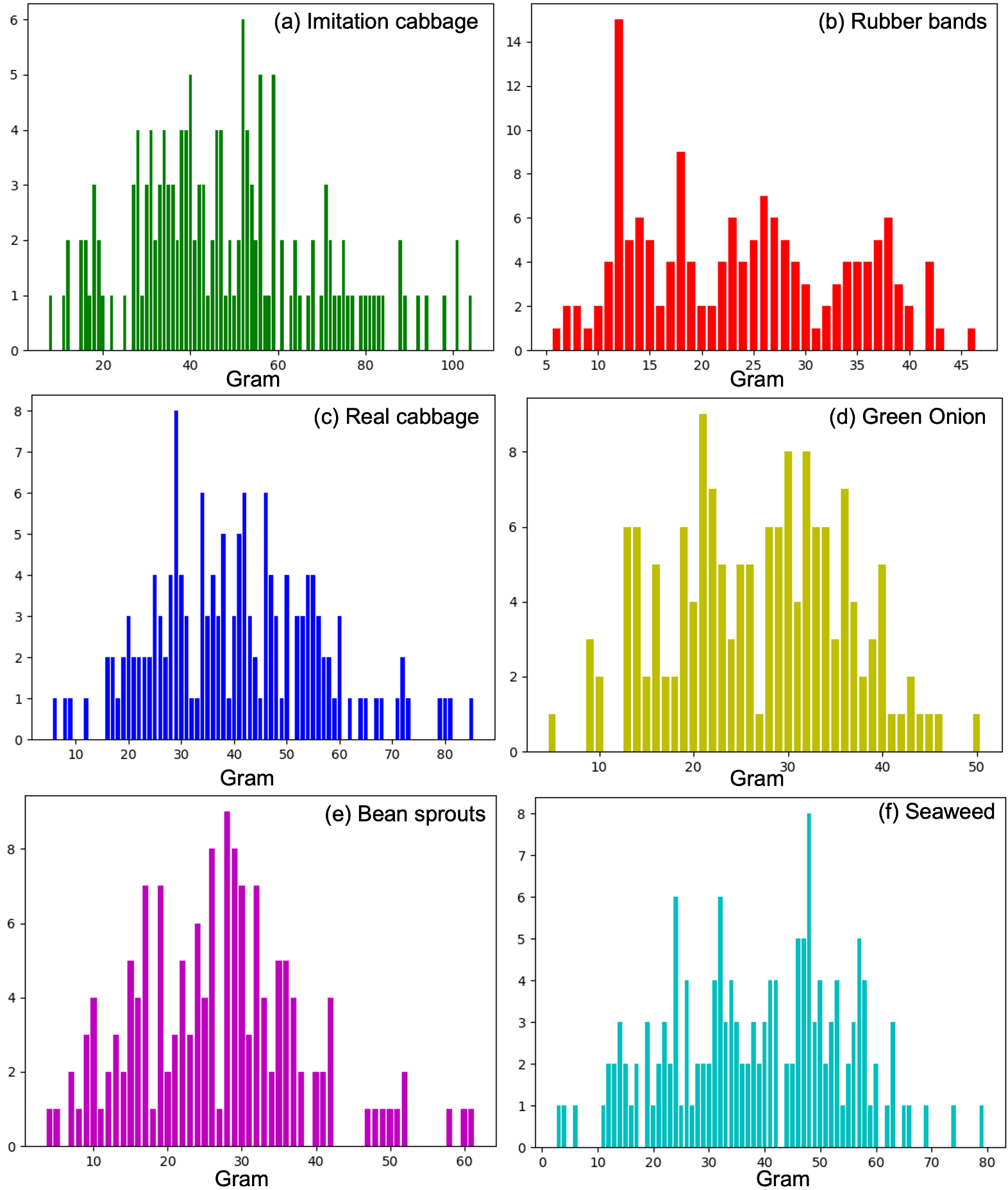}
	\caption{Histogram of masses for each food over 150 random grasps used as training data. The X-axis is the grasp mass and the Y-axis is the frequency of occurrence.}
	\label{fig:histogram}
\end{figure}
\section{Experiment Results}
\label{sec:results}
To begin with, we present the results of the data collection for all foods in Section~\ref{sec:Random Grasping for Data Collection}.
In this experiment, verification of grasp point selection, pre-grasping and post-grasping, and spines of post-grasping gripper will be evaluated in Sections~\ref{sec:Results of Grasp Point Selection}, ~\ref{sec:Results of Pre-grasping}, and~\ref{sec:Results of Post-grasping}, respectively.
In this evaluation experiment, we will use imitation cabbage and rubber bands, which are imitation foods, to evaluate the results.
They were chosen to make experimentation clear and to avoid additional confounding factors like the condition of the food changing due to experiments or time.
In addition, in Section~\ref{sec:Results Using All Methods}, all methods will be used to evaluate imitation and real foods.
\subsection{Random Grasping for Data Collection}
\label{sec:Random Grasping for Data Collection}
Figure~\ref{fig:histogram} shows the frequency with which a mass of each food was grasped over 150 grasps used as training datasets for the neural networks.
The variance of the distribution for foods is large even though there are only three grasping heights since the food surface is uneven and foods are entangled to various degrees when a grasp is executed.
Histograms of some foods clearly show multiple modes which correspond to the 3 grasp heights used.
These would suggest that heuristics that attempt to grasp a specific target amount by simply adjusting insertion depth would still lead to inaccurate, highly variable outcomes.
In the following sections, we show that we were able to achieve the target mass grasping with a high degree of accuracy by using the pre-grasping and post-grasping methods and the small number of training data (150 grasps), even though the nature of each food is different.
\begin{table}[t]
    \centering
    \caption{Success rate of grasping more than (target mass-2~g). Percentiles correspond target masses for each food with exact values provided below the food labels.}
    \label{tab:grasp point selection}
    \begingroup
    \scalefont{0.75}
        \begin{tabular}{c|c||c | c | c }
        \hline
        $\alpha$ in & Food name & $10^{th}$ & $50^{th}$ & $70^{th}$\\
        eq.~\eqref{eq:grasp_point} & (Target mass) & percentile & percentile & percentile\\
        \hline\hline
        
        $\alpha=0$ & Imitation cabbage
                          & $52\pm5$ & $80\pm4$ & $88\pm3$\\
        $\alpha=1$ & (22 g, 46 g, 56 g)
                          & $\bf{85\pm4}$ & $\bf{98\pm1}$ & $\bf{99\pm1}$\\
        \hline
                          
        $\alpha=0$ & Rubber bands
                          & $91\pm3$ & $77\pm4$ & $67\pm5$\\
        $\alpha=1$ & (12 g, 23 g, 29 g)
                          & $\bf{100\pm0}$ & $\bf{96\pm2}$ & $\bf{93\pm3}$\\
        \hline
        \end{tabular}
    \endgroup
\end{table} 
\subsection{Results of Grasp Point Selection}
\label{sec:Results of Grasp Point Selection}
In this section, we verify the method of making the grasp mass moderately larger than (target mass-2~g) by using grasp point selection for entangled foods.
To verify this, we compare the case where $\alpha=0$ in equation~\eqref{eq:grasp_point} in the grasp point selection, i.e., the same mass of grasp as the target grasp mass, and the case where $\alpha=1$ in equation~\eqref{eq:grasp_point}, i.e., more than the target grasp mass.
For each food, we chose three values for the target grasp masses, which correspond to the $10^{th}$ percentile, $50^{th}$ percentile, and $70^{th}$ percentile of masses of the 150 grasps we use as the training dataset.
When the target grasp mass is less than $10^{th}$ percentile, there are cases where the grasp amount is too small to be grasped properly.
When the value is more than $70^{th}$ percentile, we run the risk of finding no grasp point that satisfies the condition since it would need a rather large clump of food.
We evaluate the number of grasping times more than (target mass-2~g) for 100 grasps of imitation cabbage and the rubber bands.
We calculated the mean and standard deviation of success rates over the 100 grasps by bootstrapping.
The results are shown in Table~\ref{tab:grasp point selection}.

We confirm that the likelihood of grasping more than (target mass-2~g) increased by setting $\alpha=1$ for all the target grasping masses of imitation cabbage and rubber bands.
The percentage of grasps with more than the target mass does not necessarily have to be 100\%.
In practical applications in food factories, when the grasped mass is less than the target grasping mass, there is an option to re-grasp.
By comparing the time required for post-grasping, the time required for re-grasping and the rate of re-grasping, the optimal $\alpha$ for the entire system is determined.
\begin{table}[t]
    \centering
    \caption{Success rates each food that is less than or equal to $\pm$2~g for each target post-grasping with and without pre-grasping.
    }
    \label{tab:pre-grasping post-grasping}
    \begingroup
    \scalefont{0.75}
        \begin{tabular}{c|c||c c c c c}
        \hline
        \multirow{2}{*}{Pre-grasping} & \multirow{2}{*}{\shortstack{Food\\name}} & 
        \multicolumn{5}{c}{Target to drop by post-grasping}\\
        \cline{3-7}
        && 3 g & 4 g & 5 g & 10 g & 15 g\\
        \hline\hline
        w/o & \multirow{2}{*}{\shortstack{Imitation\\cabbage}}
                          & $44\pm5$ & $49\pm5$ & $40\pm7$ & $47\pm5$ & $46\pm5$ \\        
        w/ & 
                          & $\bf{87\pm3}$ & $\bf{91\pm3}$ & $\bf{92\pm3}$ & $\bf{86\pm4}$ & $\bf{88\pm3}$ \\
        \hline
        w/o & \multirow{2}{*}{\shortstack{Rubber\\bands}}
                          & $31\pm5$ & $38\pm5$ & $47\pm5$ & $51\pm5$ & $73\pm5$ \\                       
        w/ & 
                          & $\bf{94\pm2}$ & $\bf{97\pm2}$ & $\bf{95\pm2}$ & $\bf{99\pm1}$ & $\bf{95\pm2}$ \\
        \hline
        \end{tabular}
    \endgroup
\end{table}
\begin{table}[t]
    \centering
    \caption{Success rates each food with and without spines. Target mass of post-grasping is 10~g drop from the grasped amount.}
    \label{tab:spines}
    \begingroup
    \scalefont{0.75}
        \begin{tabular}{c|c||c c c c}
        \hline
        \multirow{2}{*}{Spines} & \multirow{2}{*}{\shortstack{Food\\name}} & \multicolumn{4}{c}{Success rate of diff. form the target masses}\\
        \cline{3-6}
        && w/n $\pm$2 g & w/n $\pm$3 g & w/n $\pm$4 g & w/n $\pm$5 g\\
        \hline\hline
        w/o & \multirow{2}{*}{\shortstack{Imitation\\cabbage}} 
                          & $25\pm4$ & $29\pm5$ & $32\pm5$ & $33\pm5$ \\        
        w/ & 
                          & $\bf{86\pm4}$ & $\bf{91\pm3}$ & $\bf{95\pm2}$  & $\bf{95\pm2}$\\

        \hline

        w/o & \multirow{2}{*}{\shortstack{Rubber\\bands}}
                          & $80\pm4$ & $90\pm3$ & $92\pm3$ & $99\pm1$ \\
        w/ & 
                          & $\bf{99\pm1}$ & $\bf{100\pm0}$ & $\bf{100\pm0}$  & $\bf{100\pm0}$\\
        \hline
        \end{tabular}
    \endgroup
\end{table}
\subsection{Results of Pre-grasping \& post-grasping}
\label{sec:Results of Pre-grasping}
In this section, we investigate the effect of pre-grasping and post-grasping by randomly grasping the foods with and without pre-grasping, and then from the grasped mass, 3~g, 4~g, 5~g, 10~g, and 15~g are dropped by post-grasping.
When the grasping mass is less than (target post-grasping mass+5g), re-grasping is performed.
We evaluate the percentage of grasps that are within (target post-grasping mass$\pm$2~g).
The reason for setting the target amount to $\pm$2~g is, as explained in Section~\ref{sec:Inference Process}, as its accuracy is limited by external factors of the system.
We evaluate with imitation cabbage, which tends to entangle easily, and rubber bands, which entangles less.
For each of these, a 100 trials were conducted, and the mean and standard deviation of the success rate were calculated by bootstrapping.
The results are shown in Table~\ref{tab:pre-grasping post-grasping}.
It can be seen that pre-grasping improves the accuracy of all post-grasping quantities.
The accuracy of the rubber bands, which is less entangled, is better than that of the imitation cabbage, which is more entangled, indicating that the propensity to get entangled affects the accuracy of post-grasping.
However, the difference in accuracy does not seem to depend on the amount of post-grasping after the pre-grasping.
In other words, even if the amount of post-grasping is increased by increasing the value of $\alpha$ in equation~\eqref{eq:grasp_point} in the grasp point selection, it does not affect the accuracy to achieve the target grasping mass.
\begin{table*}[t]
    \centering
    \caption{Success rates for several foods by proposal method. Numbers in bold represent better success rates than \emph{Baseline}.
    The baseline is full method without post-grasping with $\alpha=0$ in equation~(3) in grasp point selection.
    Percentiles correspond target masses for each food with exact values provided next to the food labels.}
    \label{tab:proposal method}
    \begingroup
    \scalefont{0.80}
        \begin{tabular}{c|c||c c c | c c c | c c c}
        \hline
        Method & Food name (Target masses)
        & \multicolumn{3}{c|}{$10^{th}$ percentile} & \multicolumn{3}{c|}{$50^{th}$ percentile} & \multicolumn{3}{c}{$70^{th}$ percentile}\\
        \cline{3-11}
        &(Re-grasping rates)&
           w/n $\pm$2 g & w/n $\pm$3 g & w/n $\pm$4 g & 
           w/n $\pm$2 g & w/n $\pm$3 g & w/n $\pm$4 g & 
           w/n $\pm$2 g & w/n $\pm$3 g & w/n $\pm$4 g \\
        \hline\hline
        
        Baseline & Imitation cabbage (22 g, 46 g, 56 g)
                          & $27\pm4$ & $36\pm5$ & $45\pm5$
                          & $15\pm4$ & $23\pm4$ & $33\pm5$
                          & $13\pm3$ & $18\pm4$ & $22\pm4$\\
        Ours& $(13\%, 0\%, 1\%)$
                          & $\bf{88\pm3}$ & $\bf{93\pm3}$ & $\bf{97\pm2}$ 
                          & $\bf{89\pm3}$ & $\bf{95\pm2}$ & $\bf{97\pm2}$
                          & $\bf{88\pm3}$ & $\bf{92\pm3}$ & $\bf{97\pm2}$\\
        \hline
                          
        Baseline & Rubber bands (12 g, 23 g, 29 g)
                          & $64\pm5$ & $86\pm4$ & $91\pm3$
                          & $49\pm5$ & $69\pm5$ & $80\pm4$
                          & $45\pm5$ & $56\pm5$ & $67\pm5$\\
        Ours& $(1\%, 2\%, 3\%)$
                          & $\bf{97\pm2}$ & $\bf{ 99\pm1}$ & $\bf{100\pm0}$
                          & $\bf{98\pm1}$ & $\bf{100\pm0}$ & $\bf{100\pm0}$
                          & $\bf{97\pm2}$ & $\bf{ 99\pm1}$ & $\bf{100\pm0}$\\
        \hline

        Baseline & Real cabbage (21 g, 40 g, 48 g)
                          & $10\pm4$ & $16\pm5$ & $22\pm6$
                          & $26\pm6$ & $42\pm7$ & $52\pm7$
                          & $22\pm6$ & $34\pm7$ & $44\pm7$\\
        Ours& $(6\%, 2\%, 8\%)$
                          & $\bf{88\pm5}$ & $\bf{92\pm4}$ & $\bf{96\pm3}$
                          & $\bf{88\pm5}$ & $\bf{90\pm4}$ & $\bf{92\pm4}$
                          & $\bf{88\pm5}$ & $\bf{92\pm4}$ & $\bf{92\pm4}$\\
        \hline

        Baseline & Green onion (14 g, 28 g, 32 g)
                          & $44\pm7$ & $56\pm7$ & $68\pm7$
                          & $32\pm7$ & $42\pm7$ & $52\pm7$
                          & $28\pm6$ & $38\pm7$ & $46\pm7$\\
        Ours& $(0\%, 6\%, 2\%)$
                          & $\bf{98\pm2}$ & $\bf{98\pm2}$  & $\bf{100\pm0}$
                          & $\bf{92\pm4}$ & $\bf{98\pm2}$  & $\bf{100\pm0}$
                          & $\bf{94\pm3}$ & $\bf{100\pm0}$ & $\bf{100\pm0}$\\
        \hline

        Baseline & Bean sprouts (13 g, 26 g, 32 g)
                          & $22\pm6$ & $30\pm7$ & $42\pm7$
                          & $38\pm7$ & $52\pm7$ & $60\pm7$
                          & $34\pm7$ & $50\pm7$ & $60\pm7$\\
        Ours& $(0\%, 2\%, 2\%)$
                          & $\bf{96\pm3}$ & $\bf{96\pm3}$ & $\bf{ 96\pm3}$
                          & $\bf{94\pm4}$ & $\bf{98\pm2}$ & $\bf{100\pm0}$
                          & $\bf{94\pm3}$ & $\bf{98\pm2}$ & $\bf{ 98\pm2}$\\
        \hline

        Baseline & Seaweed (17 g, 40 g, 48 g)
                          & $36\pm7$ & $46\pm7$ & $58\pm7$
                          & $24\pm6$ & $32\pm7$ & $38\pm7$
                          & $20\pm6$ & $28\pm7$ & $50\pm7$\\
        Ours& $(4\%, 0\%, 4\%)$
                          & $\bf{92\pm4}$ & $\bf{ 98\pm2}$ & $\bf{100\pm0}$
                          & $\bf{94\pm3}$ & $\bf{ 98\pm2}$ & $\bf{ 98\pm2}$
                          & $\bf{94\pm3}$ & $\bf{100\pm0}$ & $\bf{100\pm0}$\\
        \hline
        \end{tabular}
    \endgroup
    \vspace{-7.0mm}
\end{table*}
\subsection{Results of Spines of Post-grasping}
\label{sec:Results of Post-grasping}
In this section, we examine the effect of spines of post-grasping gripper.
In Section~\ref{sec:Results of Pre-grasping}, it is shown that the difference in the amount of post-grasping has no effect on the accuracy.
We let the system perform 10g post-grasping from the grasped amount with and without spines of post-grasping gripper after randomly grasping the foods with pre-grasping (See Fig.~\ref{fig:postgrasp_gripper}(b)).
When the grasping mass is less than 15g, re-grasping is performed.
Then, we evaluate the percentage of grasps that are within (target post-grasping mass $\pm$2~g, $\pm$3~g, $\pm$4~g, and $\pm$5~g) for 100 grasps of imitation cabbage and rubber bands.
We calculated the mean and standard deviation of success rate for the 100 grasps by bootstrapping.
The results are shown in Table~\ref{tab:spines}.

It can be seen that the presence of spines allows for accurate post-grasping of both imitation cabbage and rubber bands.
Since imitation cabbage does not reach 100\% success even within $\pm$5~g, it uncontrollably falls off in clumps for foods with severe entangles.
Without the spines, the accuracy is extremely poor.
In the case of no spines, the food tends to fall out in uncontrollable clumps or, the worst case, all the food would fall out at once.
This indicates that the presence of the spines enables post-grasping to dispose of a small amount of food at a time while carefully untangling the food.
\subsection{Results Using All Methods}
\label{sec:Results Using All Methods}
In this section, we evaluate the proposed methods for grasp point selection, pre-grasping and post-grasping, when used together.
The baseline used is when grasping that is done according to the predicted mass with $\alpha=0$ in equation~\eqref{eq:grasp_point} in grasp point selection with pre-grasping without post-grasping.
We use imitation cabbage and rubber bands as imitation foods.
In addition, we use numerous real foods including shredded cabbage, sliced green onion, bean sprouts, and shredded seaweed (See Fig.~\ref{fig:foods}).
The target grasp masses are the $10^{th}$, $50^{th}$, and $70^{th}$ percentile of grasped masses over the 150 grasps collected and used as the training dataset (see Table~\ref{tab:proposal method} for specific values).
We apply the proposed method and perform a 100 grasps for imitation foods and 50 grasps for real foods.
Grasps which fail to reach a mass greater than the (target-2~g) are re-grasped.
We show re-grasping rate for each target mass provided below the food labels (see Table~\ref{tab:proposal method}).
We report success rates when the grasped mass is within (target mass $\pm$2~g, $\pm$3~g, and $\pm$4~g). 
Baseline is 100 grasps for imitation foods and 50 grasps for real foods. 
For each food, the mean and standard deviation of success rates were calculated by bootstrapping over a 100 grasps for imitation foods and 50 grasps for real foods.
Since the actual food spoil and change in profile over the duration of the experiment due to grasping, temperature, loss of water, etc, we limit the number of grasps to 50.
The results are shown in Table~\ref{tab:proposal method}.

It can be seen that the proposed method significantly outperforms the baseline across all experiments.
The results show that the accuracy of the proposed method is remarkably high (88\%) for within $\pm$2~g thereby proving its efficacy.
Furthermore, the proposed method has the same accuracy independent of the target masses.
The system works even if the amount of food in the container is gradually reduced (See attached video).
The time required for post-grasping was roughly between 2--10 seconds.
The percentage of re-grasping tended to be higher in the case of foods that were easily entangled, such as imitation and real cabbage.

For practical use, it is necessary to adjust the value of $\alpha$ in equation~\eqref{eq:grasp_point} in the grasp point selection based on the time required for post-grasping, re-grasping and the likelihood of needing re-grasping, as described in Section~\ref{sec:Results of Grasp Point Selection}.
In addition, to accommodate time/accuracy requirements, we can omit pre-grasping and post-grasping when it is desired to shorten the pick-and-place time at the cost of reduced accuracy.
In addition, although we used an expensive depth sensor with high accuracy, it may be possible to use a less accurate but cheaper one by increasing the magnitude of $\alpha$ to compensate for the loss of accuracy in grasping amount estimation.
\section{Conclusion}
\label{sec:conclusion}
In this paper, we propose three methods for grasping the target mass of entangled food:
1) grasping more than the target mass considering the uncertainty,
2) pre-grasping to reduce the degree of the entanglement of food, and
3) post-grasping to adjust the grasping mass when the grasped mass is more than the target mass.
For each of the three methods, we confirmed a significant improvement in accuracy by evaluating them with imitation food.
In addition, we combined all three methods and evaluated them with imitation foods and real foods.
If we simply predict the grasp amount as a baseline, most of the foods do not reach 50\%.
In contrast, the proposed method achieves an accuracy of over 88\% for foods we evaluated.

\bibliographystyle{IEEEtran} 
\bibliography{IEEEabrv,bibliography}
\end{document}